\def\year{2021}\relax
\documentclass[letterpaper]{article} 
\usepackage{aaai21}  
\usepackage{times}  
\usepackage{helvet} 
\usepackage{courier}  
\usepackage[hyphens]{url}  
\usepackage{graphicx} 
\usepackage{amsmath}
\usepackage{subfigure}
\usepackage{multirow}
\usepackage[switch]{lineno}  
\usepackage{lipsum}
\usepackage{natbib}  
\usepackage{caption} 
\title{Denoising Distantly Supervised Named Entity Recognition via a Hypergeometric Probabilistic Model}

\author{
	Wenkai Zhang\textsuperscript{\rm 1,3,*},
	Hongyu Lin\textsuperscript{\rm 1,*}, 
	Xianpei Han\textsuperscript{\rm 1,2,$\dagger$},
	Le Sun\textsuperscript{\rm 1,2,$\dagger$},
	Huidan Liu\textsuperscript{\rm 1}, \\
	Zhicheng Wei\textsuperscript{\rm 4},
	Nicholas Jing Yuan\textsuperscript{\rm 4} \\
}

\affiliations{
	\textsuperscript{\rm 1}Chinese Information Processing Laboratory ~ 
	\textsuperscript{\rm 2}State Key Laboratory of Computer Science \\
	Institute of Software, Chinese Academy of Sciences, Beijing, China\\
	\textsuperscript{\rm 3} University of Chinese Academy of Sciences, Beijing, China \\
	\textsuperscript{\rm 4}Huawei Cloud\&AI \\
	{\{wenkai2019,hongyu,xianpei,sunle,huidan\}@iscas.ac.cn} \\
	{\{weizhicheng1,nicholas.yuan\}@huawei.com}\\
}

\begin{document}

\maketitle

\newcommand\blfootnote[1]{%
\begingroup
\renewcommand\thefootnote{}\footnote{#1}%
\addtocounter{footnote}{-1}%
\endgroup
}

\begin{abstract}
Denoising is the essential step for distant supervision based named entity recognition. Previous denoising methods are mostly based on instance-level confidence statistics, which ignore the variety of the underlying noise distribution on different datasets and entity types. This makes them difficult to be adapted to high noise rate settings. In this paper, we propose \emph{Hypergeometric Learning (HGL)}, a denoising algorithm for distantly supervised NER that takes both noise distribution and instance-level confidence into consideration. Specifically, during neural network training, we naturally model the noise samples in each batch following a hypergeometric distribution parameterized by the noise-rate. Then each instance in the batch is regarded as either correct or noisy one according to its label confidence derived from previous training step, as well as the noise distribution in this sampled batch. Experiments show that HGL can effectively denoise the weakly-labeled data retrieved from distant supervision, and therefore results in significant improvements on the trained models.
\blfootnote{\hspace{-0.3cm}$\dagger$: Corresponding authors.}
\blfootnote{\hspace{-0.3cm}*: Equal Contribution. H. Lin designed the method and wrote the paper. W. Zhang conducted the experiments and detailed analysis.}
\end{abstract}

\section{Introduction}
Named Entity Recognition (NER), aiming to identify text spans pertaining to specific semantic types such as person, organization and location, is a foundational NLP task. In recent years, supervised neural network-based approaches ~\cite{lample-etal-2016-neural,chiu-nichols-2016-named,ma-hovy-2016-end} , which can automatically extract underlying features from annotated texts and conduct NER recognition, have achieved promising results in almost all NER benchmarks.

\begin{figure}[!ht]
	\centering
	\setlength{\belowcaptionskip}{-0.5cm}
	\subfigure[Same setting, differnet types]{
		\includegraphics[width=0.223\textwidth]{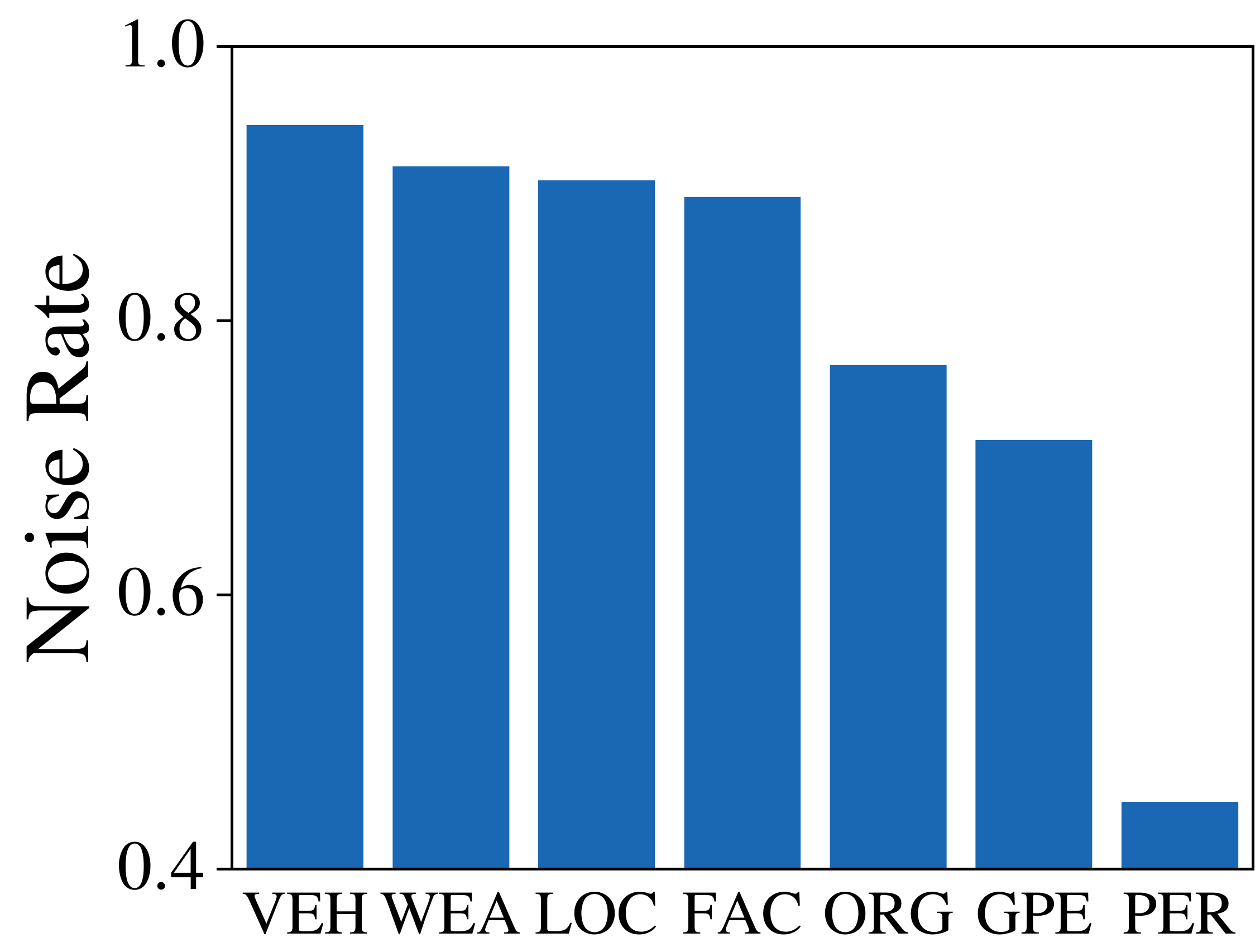}
	}
	\subfigure[Same type, different settings]{
		\includegraphics[width=0.223\textwidth]{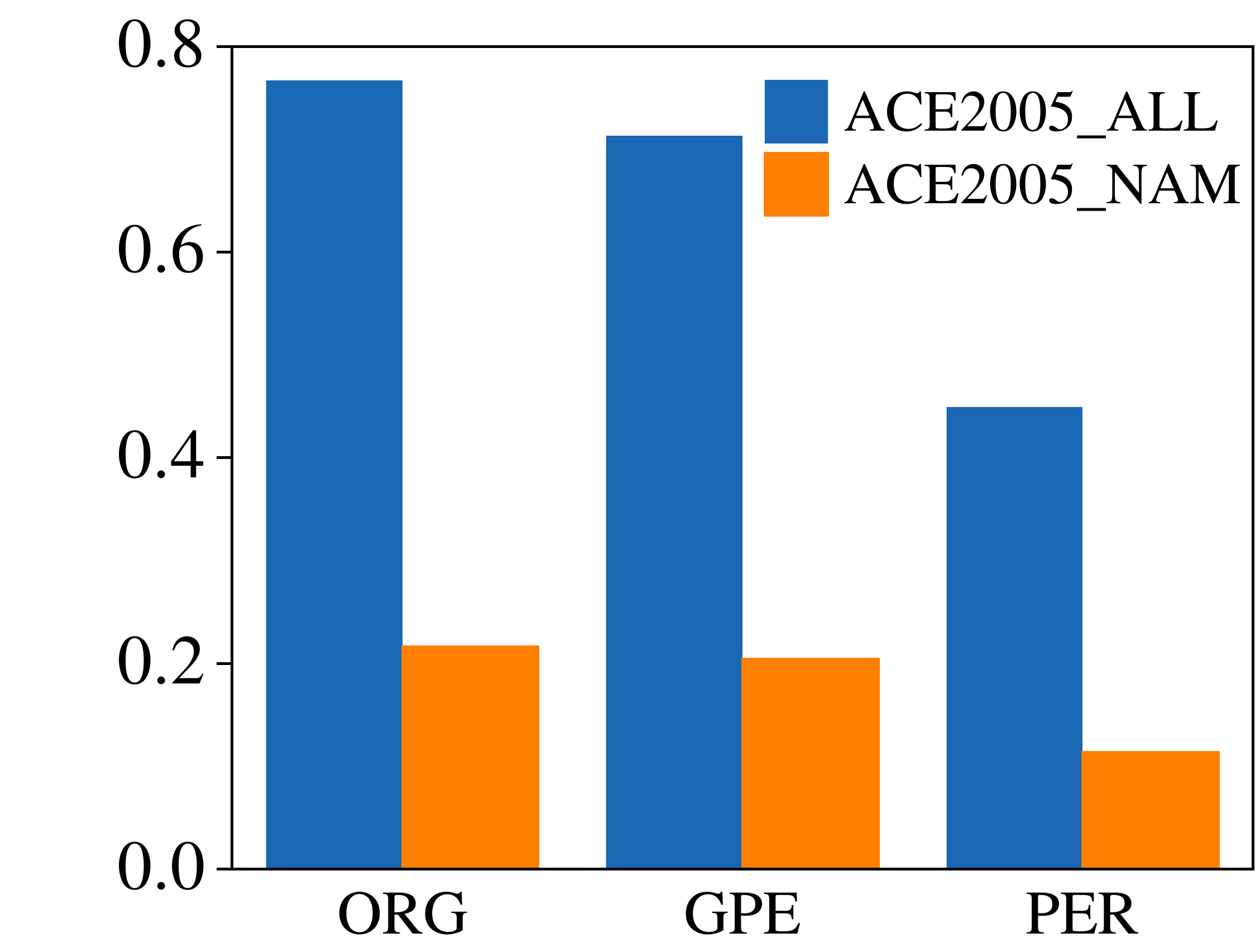}
	}
	\caption{Precision-Recall curves of different denoising algorithms on ACE2005 datasets. We can see that HGL significantly outperforms other baselines, especially on high noise rate ACE\_ALL setting.}
	\label{fig:noiserate}
\end{figure}

Even with great success, supervised learning methods heavily rely on fully-annotated training data. However, annotated training data is too expensive to obtain, which restricts the real-world application of current NER models in various entity types. To tackle this problem, distant supervision based NER (DS-NER)~\cite{renxiang2017,yang-etal-2018-distantly,shang-etal-2018-learning,peng-etal-2019-distantly,nooralahzadeh-etal-2019-reinforcement} methods are introduced. Generally, DS-NER uses easily-available resources (commonly dictionaries in NER) to automatically label plain texts to produce large-scale data, which is generally referred as \emph{weakly annotated data}. And then the generated training data will be used to learn neural network models. For example, by matching person name ``Washington'' in dictionary with text ``Washington is the first U.S. president'', DS-NER will produce a training instances with entity type PERSON. As unlabeled plain texts and entity mention dictionaries are very easy to obtain, distant supervision based methods can significantly reduce the annotation efforts and produce much larger scale of annotation data for model learning.

\begin{figure*}[!ht]
	\centering
	\includegraphics[width=0.65\linewidth]{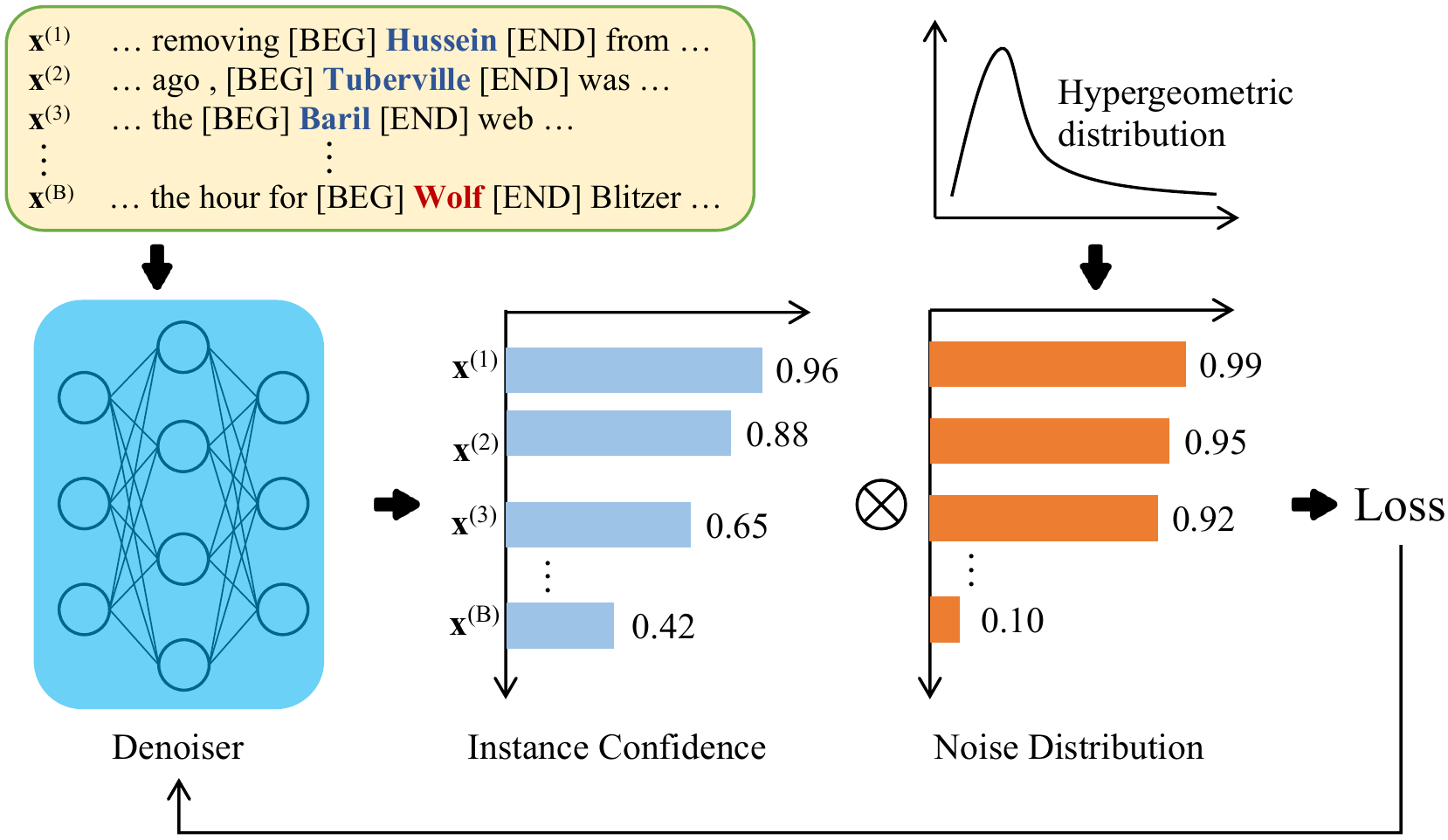}
	\caption{The overall architecture of HGL. At each step, a batch of instances is sent into a neural network-based denoiser to obtain its correct/noisy confidence. Then the confidence of each instance is incorporated with the overall hypergeometric noise distribution of this batch to obtain the training loss of current batch.}
	\label{fig:denoiser}
\end{figure*}

Unfortunately, although DS-NER boosts the size of training data, it still faces critical high noise challenge. Due to the ambiguity of natural language, distant supervision will inevitably introduce remarkable noise into weakly labeled data. For example, ``Washington'' can refer to either a person name or a location name and DS will generate wrong PERSON instances such as ``Washington, capital of the United States''. As a result, denoising is critical to achieve promising DS-NER performance. Currently, most DS-NER denoising methods are based on instance-level confidence statistics, which predict whether one instance is noise according to its feature similarity to the majority of other instances. These instance-level denoising algorithms, unfortunately, ignore the underlying noise structure of different datasets. For example, the dictionary qualities and dataset properties vary significantly in different DS settings, which will seriously affect the denoising decisions. 
To illustrate this, Figure~\ref{fig:noiserate} shows the noise rate of DS-NER on ACE2005 datasets. We can see that the noise rates on weakly-labeled training data vary significantly from different entity types and different settings (e.g., named mention only v.s. all mentions). In some settings, even a great majority of weakly-labeled instances are noise. Consequently, conventional instance-level methods cannot achieve robust performance on different noise DS-NER settings adaptively, and therefore it is critical to take the noise distribution into consideration.

To this end, this paper proposes \emph{\textbf{Hypergeometric Learning (HGL)}}, which can effectively and adaptively denoise DS-NER by further modeling and leveraging the underlying noise structure of a dataset. HGL learns to denoise distantly supervised labeled data by reshaping conventional training procedure according to a noise rate-aware hypergeometric probabilistic model.  Figure~\ref{fig:denoiser} shows the overall framework of HGL. Specifically, to model the noise distribution during neural network training, we formulate each mini-batch of size $B$ as a subset being drawn from a weakly labeled data of size $N$ with noise rate $p$. Then the noise sample size $S$ in this mini-batch will naturally follow a hypergeometric distribution, i.e., $S \sim H(N,N \times p,B)$.  Therefore, we regard each instance as either correct or noisy one according to its label confidence derived from previous training step, as well as the noise distribution in the sampled batch. For example, if one training batch is expected to have 5 instances to be correctly labeled while the others to be noisy, then the instances with top-5 confidence will be regarded as correct instances, while others are treated as noisy ones. 

Based on the above principles, we propose a neural network-based denoiser for DS-NER. The training loss of the denoiser is marginalized over the hypergeometric distribution $S$ to consider all potential noise distribution of the current training batch. For different DS-NER settings with different noise rates, the hypergeometric distribution can be easily adapted by only adjusting the noise rate according to the reality. Compared with traditional instance confidence-based denoising methods, the proposed HGL explicitly exploits the noise distribution during the denoising procedure. This enables more robust performance among different NER settings, especially when the noise rate is relatively high.

Generally, the main contributions of this paper are:

1) \textbf{We propose to naturally model the noise distribution of DS-NER during training using a hypergeometric probabilistic model.} The model exploits the underlying noise structures, and therefore can be more robustly adapted to different DS-NER Settings. To the best of our knowledge, this is the first work which tries to consider this critical information in DS-NER denoising.

2) \textbf{Based on the noise model, we propose Hypergeometric Learning (HGL), a denoising algorithm for DS-NER. } By reshaping the training procedure according to the noise distribution and instance-level confidence, HGL can robustly denoise the training data and therefore leads to better NER models. Because denoising is a critical step for all distant supervision settings, we believe our idea and method can potentially benefit many other DS tasks, such as relation extraction.

3) We conducted experiments under various distant supervision settings on ACE2005 and CoNLL03 NER datasets. Experimental results show that HGL can effectively denoise DS-NER data. Furthermore, using the datasets denoised by HGL to train a novel NER model can also significantly boost the system performance, compared with using only weakly-labeled data and other denoising algorithms.

\section{Related Work}
Neural network-based models have achieved very promising results in NER field~\cite{lample-etal-2016-neural,chiu-nichols-2016-named,ma-hovy-2016-end,lin-etal-2020-rigorous}. These methods commonly rely on large-scale training data to learn effective entity mention recognizer. However, fully-annotated training data is too expensive to obtain, which limits the application of these methods to more NER scenario.

Distant supervision~\cite{mintz-etal-2009-distant}, aiming to efficiently generate large-scale training data using easily available resources, has attracted much attention~\cite{zeng-etal-2015-distant,lin-etal-2016-neural}. For NER, one common practice is to use an entity mention dictionary to automatically annotate the plain texts to generate large-scale training data~\cite{renxiang2017,yang-etal-2018-distantly,shang-etal-2018-learning,peng-etal-2019-distantly,nooralahzadeh-etal-2019-reinforcement}. Unfortunately, DS-NER will introduce remarkable noise into the training data, and therefore undermines the NER performance.

To tackle this problem, one solution is to only use a high precision dictionary to avoid introducing noise, and then denoising algorithms are applied to handle false negative instances~\cite{yang-etal-2018-distantly,shang-etal-2018-learning,peng-etal-2019-distantly,nooralahzadeh-etal-2019-reinforcement}. These approaches have shown promising results in some specific domains such as medical NER. However, due to the ambiguity of natural language expressions, it is often impossible to obtain a high precision dictionary for many domains. For example, there are many location names (e.g., Washington) which could also be used as person names. This limits the application of these methods to more general NER tasks.

Instead of assuming using a high precision dictionary, this paper proposes Hypergeometric Learning, which  can be directly applied to various distant supervision settings. Furthermore, we also proposed a \emph{Mention Blocking} approach to extend our framework into the circumstance where extreme low false negative rate exists.
There are some previous studies along this direction, but the majority of them are based on instance-level confidence derived from an Expectation-Maximization style procedure~\cite{moon1996expectation}. These methods mainly reduce the impact of noise on the training procedure by rescaling the instances according to their confidence. However, they did not explicitly model the noise distribution during the neural network training, and therefore unable to achieve robust performance when high noise rate exists.

\section{Hypergeometric Learning for Denoising Distantly Supervised NER}
As we mentioned above, the quality of corpus and dictionaries varies significantly from different DS-NER settings. As a result, denoising methods merely based on instance-level confidence to identify noise can not achieve promising results, because in many circumstance even the majority of the datasets would be mislabeled instances. Instead, this paper proposes to address this issue by jointly considering the underlying noise distribution and the instance-level confidence. The essential of our method is to learn a neural network-based denoiser using Hypergeometric Learning (HGL) algorithm. In the following, we will first illustrate the structure of our denoiser, then introduce our Hypergeometric Learning algorithm.

\subsection{Denoiser}
Our denoiser is a neural network model that takes the weakly annotated instances from distant supervision as input and predicts whether these instances are correctly labeled entity mentions. 
Specifically, given an input sentence $\{x_1,x_2,...,x_n\}$ and a weakly labeled candidate mention $\{x_i,x_{i+1},...,x_j\}$ in it, we follow~\citet{baldini-soares-etal-2019-matching} to represent an instance $x$ by adding a start symbol [BEG] before $x_i$ and an end symbol [END] after $x_j$:
\begin{equation}
\mathbf{x} = [x_0,...,[BEG],x_i,...,x_j,[END],...,x_{n}].
\end{equation}
Then $\mathbf{x}$ is sent into a Transformer~\cite{transformer} pretrained using BERT~\cite{devlin-etal-2019-bert}. We use $\mathbf{h_i}$ to denote the encoder output of token $x_i$, and $\mathbf{h_{BEG}}$ and $\mathbf{h_{END}}$ to denote the representations at token [BEG] and [END] respectively. Then we apply an attention-based mechanism to obtain the representation of the candidate mention $\mathbf{r}$:
\begin{equation}
\mathbf{r} = \sum_{k = [BEG]}^{[END]} {\alpha_k \mathbf{h}_k},
\end{equation}
where $\alpha_k$ can be regarded as an important score of token $x_k$ to the entire instance, which is computed by
\begin{equation}
\alpha_k = \frac{\exp(\mathbf{Wh_i} + \mathbf{b})} {\sum_{k = [BEG]}^{[END]} \exp(\mathbf{Wh_k} + \mathbf{b})} .
\end{equation}
The representation $\mathbf{r}$ is then sent into a multi-layer perceptron followed by a sigmoid layer to predict the probability $f(x)$ of this instance being a correctly labeled  mention:
\begin{equation}
\label{eq:conf}
f(x) = \sigma(MLP(\mathbf{r})),
\end{equation}
where $\sigma$ is the sigmoid function and MLP is multi-layer perceptron without active function at the top. The denoiser is expected to output a larger $f(x)$ (i.e., more close to 1) if the instance is more likely to be a correctly labeled instance (i.e., an correct entity mention), while output a smaller $f(x)$ (i.e., more close to 0) if this instance is a noisy one.

\subsection{Hypergeometric Learning for Denoiser Training}
Commonly, neural network-based models are trained in a supervised learning paradigm using batch-wise learning algorithm. However, due to the lack of correct/noisy annotation data, it is impractical to directly train our denoiser using conventional neural network learning criteria.

\paragraph{Hypergeometric Probabilistic Model.} 
In this paper, instead of using annotated data, we propose to tackle this problem by exploiting the noise distribution during denoiser training. Specifically, if there are $N$ weakly-labeled instances in total and the accuracy of this dataset is expected to be $p$ (i.e., the noise rate is $1-p$). During neural network training, a common practice is to sample $B$ instances from the overall $N$ population. Then the number of correctly labeled entity mentions $S$ in each batch will naturally follow the hypergeometric distribution:
\begin{equation}
S \sim  H(N,K,B),
\end{equation}
where $K=N \times p$ is the expected size of correctly labeled instances and $H$ is the hypergeometric distribution, which can be computed as:
\begin{equation}
P(S=k) = \frac{\binom{K}{k} \binom{N-K}{B-k}}{\binom{N}{B}},
\end{equation}
where $\binom{A}{B} =\frac{A!}{B!(A-B)!}$. For clarity, we will use $Q_k$ to denote $P(S=k)$ in the rest of this paper.

\paragraph{Hypergeometric Learning for DS-NER.} 

Once we know the distribution of correct/noise sample size in each batch, we will evaluate the potential of each instance being a correct or noisy one according to its label confidence derived from previous training step as well as the noise distribution in the current batch. For example, if in this batch there are only 5 instances expected to be the correct instances according to the hypergeometric distribution, we will regard instances with top-5 confidence (i.e., $f(x; \theta)$ in Equation~\ref{eq:conf}) as the entity mentions, while other instances in this batch will be regarded as noisy instances. The loss function during training will be marginalized over the hypergeometric distribution to consider all potential noise distribution.

Formally, let \{$\mathbf{x^{(1)}},\mathbf{x^{(2)}},...,\mathbf{x^{(B)}}$\} denote the instances in the current sampled batch in a descending order according to their correct confidence in the previous step. Our Hypergeometric Learning will optimize the denoiser in the current training set according to the following loss function:
\begin{equation}
L(\theta) = \sum_{i=1}^{B} \omega_i \log f(x^{(i)}) + (1-\omega_i) \log(1 - f(x^{(i)})),
\end{equation}
where $f(x)$ and $1-f(x)$ are the denoiser output of the probability of instance $x$ being a correct or noisy instance respectively. $\omega_i$ is the weight that indicates instance $x^{(i)}$ should be regarded as a correct entity mention or a noisy instance, which is derived from the hypergeometric distribution by:
\begin{equation}
\omega_i = \sum_{k=i}^{B} Q_k.
\end{equation}
This equation indicates that if an instance ranks at $k^{th}$ according to the confidence, it will be regarded as a correct instance only if at least $k$ instances in this batch are expected to be correct instances. Therefore, HGL is able to exploit both the instance-level confidence and the overall noise distribution in a unified training procedure, and thus can achieve robust performance on various settings only by adaptively adjusting the expected noise rate according to the underlying noise distributions.

\subsection{Dealing with Extremely Low False Negative Rate with Mention Blocking}
As described above, the Hypergeometric Learning mainly focuses on the circumstance where the dictionary is with high recall rate. However, in practice the applied dictionary commonly cannot cover all mentions in the dataset, and therefore requires the model to also deal with false negative instances. Considering the fact that entity mentions of a specific type only cover a very minority of all noun phrases (commonly less than 1\%), it is quite difficult to distinguish false negative instances from the true negative ones using standard denoising techniques. To tackle this problem, we further propose a \emph{Mention Blocking} approach, which first screens possible entity mentions of a specific type by leveraging the mention compositional knowledge learned from the dictionary, then the potential mentions will be further introduced into Hypergeometric Learning procedure, so that false negatives can also be denoised.

Specifically, we regard all noun phrases in the entire corpus (except mentions matched by the distant supervision dictionary) as the potential mentions. Then inspired by~\citet{lin-etal-2019-gazetteer}, we learn a Bert-based classifier from given dictionaries, which takes a context-free noun phrase (mention utterance) as the input, and outputs a score indicating how possible the input phrase could be a valid mention of such entity type. For example, assigning ``Washington'' a high PER score and ``New York'' a low PER score. After that, we collect the noun phrases with high classification scores as the possible entity mention block. And only the instances within the block will be regarded as potential false negative mentions. Then Hypergeometric Learning will be applied to denoise these potential mentions, which tries to further distinguish false negative instances (i.e., the noises in true negative instances) from all noun phrases screened out. The learning procedure regards the false negative instances as the noise out of all blocked noun phrases. And the denoiser is learned jointly from denoising both positive and negative instances, which means that we will learn one denoiser with the loss from both these blocked noun phrases and the instances from distant supervision.

\section{Experiments with Synthetic Dictionaries}
In this section, we conducted experiments to verify the efficacy of the proposed Hypergeometric Learning algorithm. Specifically, we evaluate our method to answer the following three questions:
\begin{itemize} 
	\item Can Hypergeometric Learning-based denoiser effectively identify the noisy instances in the weakly-labeled training data?
	\item Can the proposed Mention Blocking approach effectively detect false negative instances and improves the denoising performance?
	\item Can using the dataset denoised by HGL significantly improve the performance of conventional NER models such as Bert-CRF tagger?
\end{itemize}

\subsection{Experimental Settings}

\paragraph{Dataset.}
We conduct experiments on ACE2005~\cite{walker2006ace} benchmark, which contains 7 entity categories.
We use the same experimental setup as ~\cite{ju2018neural,katiyar2018nested,wang2018neural,strakova2019neural,lin-etal-2019-sequence}, where the entire dataset is split into 8:1:1 for training, developing and testing respectively. We keep only the outmost entity mentions and therefore ignore the overlapping mentions of the same type.
To verify the adaptiveness of the proposed methods, we conducted experiments based on two settings: 1) Vanilla ACE2005, which considers all named, nominal and pronominal mentions; 2) ACE2005 NAM\footnote{For ACE2005 NAM, we only conduct experiments on PER, ORG and GPE because other entity types do not have sufficient instances in both training and test set.}, which considers only named mentions in the original dataset, while ignores nominal and pronominal mentions. We refer to these two settings as ACE\_ALL and ACE\_NAM respectively. Generally speaking, ACE\_ALL is with relatively large noise ratio compared with ACE\_NAM due to the stronger ambiguity of nominal and pronominal utterances.

\paragraph{Distant Supervision.} To effectively verify the adaptiveness of Hypergeometric Learning under different dictionary quality settings, in this section we conducted experiments using synthetic dictionaries extracted from the training data. And in the next section we will conduct an experiment on the real-world dictionaries to further verify the effectiveness of HGL. 

Specifically, in this section we used all mentions extracted from the original training data as our distant supervision source dictionary. We then used the dictionary to label all documents in the original training set to obtain weakly-labeled training instance for our denoiser. This guarantees that the dictionary used in our experiment is with high recall rate. The macro-averaged noise rate on ACE\_ALL and ACE\_NAM are 0.79 and 0.20, respectively. This indicates the necessity of introducing denoising algorithm to distant supervision NER, even when a golden dictionary is applied. 

\paragraph{Noise Rate Estimation.} For HGL, we estimate the noise ratio of each entity type from the development set. Because we may not estimate noises rate exactly accurately in reality, we approximate noise rates with its nearest 5\% percentile. For example, if the noise rate is 34.1\% in the development set, we will use 35\% as the noise rate during HGL learning.

\paragraph{Implementation Detail.}

We use Allennlp ~\cite{gardner-etal-2018-allennlp}, an open-source NLP research library to implement our method and conduct experiments. For HGL, we set batch size to 150 and use Adam ~\cite{Adam} as optimizer, the learning rate is $1\times10^{-5}$. We used \emph{cased\_L-24\_H-1024\_A-16} as pretrained encoder. We openly release our source code at {github.com/zwkatgithub/HGL}.

\subsection{Baselines}
We compared HGL with the following baselines:
\begin{itemize} 
	\item \textbf{Supervised Learning (Supervised)}, which used fully-annotated data to directly train the NER model, which can serve as the upper bound of all denoising methods.
	
	\item \textbf{Naive Distant Supervision (Naive)}, which directly used the dictionary to label the plain texts and does not conduct any denoising on the annotation results.
	
	\item \textbf{Instance-level Expectation Maximization (Instance)~\cite{moon1996expectation}}, which identifies noisy training instances in a self-supervision manner according to the instance-level confidence. Specifically, the loss of each instance is reshaped by:
	\begin{equation}
	\begin{aligned}
	L(x) = & f(x;\theta_t) \log f(x) \\
	&+ [ 1-f(x;\theta_t) ] \log [1-f(x)].
	\end{aligned}
	\end{equation}
	This can be regarded as a baseline that only considers instance-level confidence but ignores the overall noise distribution during the training procedure.
	
	\item \textbf{XR-Loss~\cite{xr} (XR)}, which considers the proportional mapping between distant supervision label mentions to correctly labeled and noisy instances. In other words, this method takes only the overall noise rate into consideration but ignores the instance-level confidence.
	
	\item \textbf{Positive-Unlabeled Learning~\cite{peng-etal-2019-distantly} (PU-Learning)}, which tackles the mislabeling in distant supervision by regarding it as a positive-unlabeled learning problem and reshapes the training procedure with a regularizer considers the mislabeling ratio. Note that this method is to learn a NER model directly on DS-NER datasets, which is unable to identify the exact noise instance during learning.
\end{itemize}

\begin{table*}[!ht]
	\centering
	\setlength{\belowcaptionskip}{-0.3cm}
	\begin{tabular}{lrrrrrrr|rrr}
		\hline
		Dataset & \multicolumn{7}{c|}{ACE\_ALL}             & \multicolumn{3}{c}{ACE\_NAM} \\
		\hline	
		Method  & PER & ORG & GPE & LOC & FAC & VEH & WEA & PER       & ORG       & GPE       \\
		\hline
		Instance    &  63.97   &  72.88   &  66.15   &  83.75   &  77.61   &  79.23   &  80.92   &     71.65      &     50.07      &     52.41      \\
		XR      &  56.51   &  77.42   &  67.35   &  82.03   &  78.62   & 84.63   &  74.47   &     66.97      &     55.30      &     59.46      \\
		HGL   &  \textbf{80.35}   &  \textbf{82.28}   &  \textbf{86.50}   &  \textbf{93.86}   &  \textbf{90.56}   &  \textbf{90.03}   &  \textbf{94.00}   &     \textbf{71.68}      &     \textbf{59.22}      &     \textbf{64.35}      \\
		\hline
	\end{tabular}
	\caption{The AUC scores of denoising performance of different algorithms on ACE\_ALL and ACE\_NAM. We can see that HGL significantly outperfoms other methods by a large margin.}
	\label{tab:auc}
\end{table*}

\subsection{Denoising Performance}
The first group of our experiments was conducted to evaluate how well the proposed HGL and baseline methods can denoise the weakly labeled data. We evaluate the denoising performance using ranking based metrics to see how well the algorithms can identify correct instances out of noisy instances.

\begin{figure}[!ht]
	\centering
	\setlength{\belowcaptionskip}{-0.5cm}
	\subfigure[ACE\_ALL]{
		\includegraphics[width=0.22\textwidth]{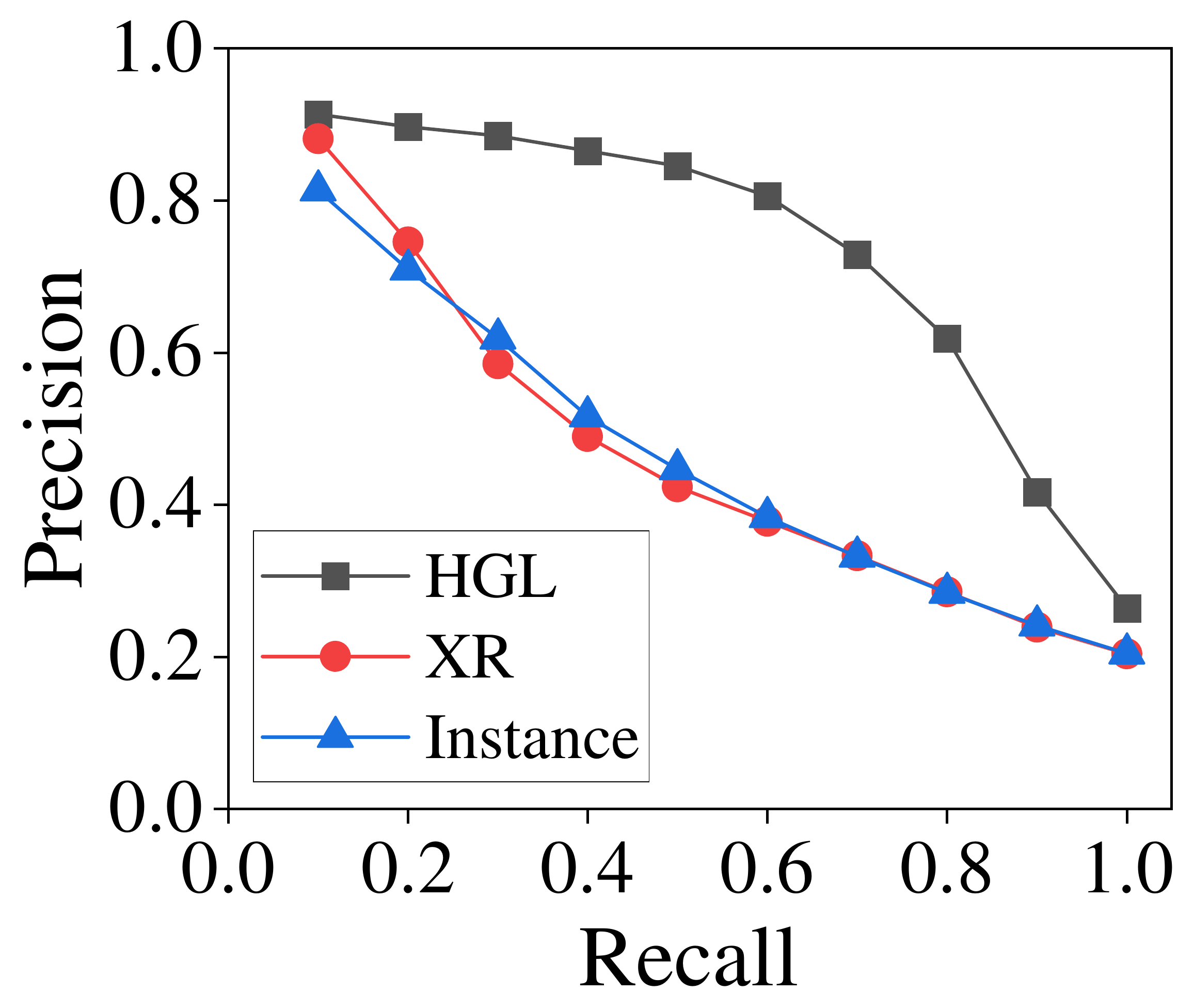}
		\label{ace2005_map_per}

	}
	\subfigure[ACE\_NAM]{
		\includegraphics[width=0.22\textwidth]{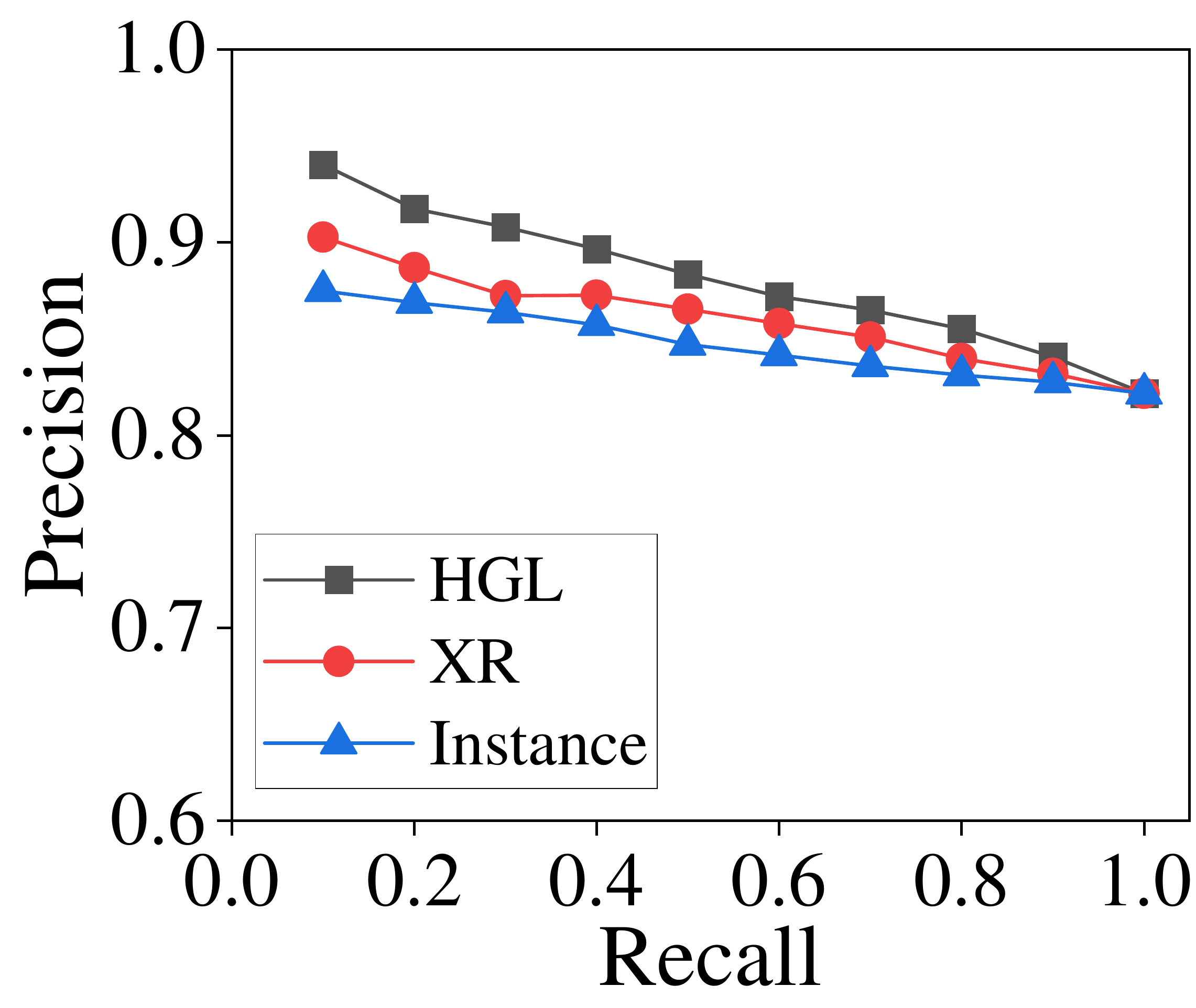}
		\label{ace2005name_map_per}
	}
	\caption{Precision-Recall curves of different denoising algorithms on ACE2005 datasets. We can see that HGL significantly outperforms other baselines, especially on high noise rate ACE\_ALL setting.}
	\label{fig:map}
\end{figure}

Figure~\ref{fig:map} shows the macro-averaged precision-recall curves of HGL and other baselines on ACE\_ALL and ACE\_NAM respectively, while Table~\ref{tab:auc} shows the overall AUC scores of all systems. 
We can see that the proposed HGL algorithm significantly outperforms other denoising algorithms. For AUC scores, HGL outperforms other baselines by a large margin on almost all settings, which demonstrates the effectiveness of our method. Furthermore, we can see that from the Precision-Recall curves, the performance of other baselines dramatically drops as the recall increases, which indicates that HGL performs significantly better, especially on correct instances that are hard to be distinguished from noise. Besides, from both the AUC scores and the Precision-Recall curves, the improvement of HGL, compared with other baselines, is more remarkable on ACE\_ALL than on ACE\_NAM. This is because, as we mentioned above, the overall noise rate of ACE\_ALL is much larger than ACE\_NAM, which makes previous methods that do not consider both instance confidence and noise distribution can not achieve promising performance in this circumstance. Generally speaking, owing to the effective exploit both instance-level confidence and the overall noise distribution, HGL can achieve much better and robust denoising performance for distant supervision NER.

\begin{table*}
	\centering
	\begin{tabular}{l|rrrrrrr|rrr}
		\hline
		& \multicolumn{7}{c|}{ACE\_ALL} & \multicolumn{3}{c}{ACE\_NAM} \\
		\cline{2-11}
		Model & PER & ORG & GPE & LOC & FAC & VEH & WEA  & PER & ORG & GPE \\
		\hline\hline
		Supervised 	&  85.51   &  66.45   &  76.92   &  65.96   &  58.97   &  52.29   &  73.12 & 84.80 & 76.55 & 89.89      \\ \hline
		Naive	&  60.78   &  32.70   &  33.28   &   7.91   &  15.22   &   8.12   &   7.47   & 81.03 & \textbf{70.36} & 75.63      \\
		Instance    &  54.52   &  32.53   &  34.43   &  30.99   &  28.07   &  11.32   &  18.75   & 82.84 & 51.83 & 76.53         \\
		XR      &  50.60   &  38.10   &  28.38   &  22.54   &  27.43   &  35.21   &  14.46   & 80.12 & 57.92 & 71.40     \\
		PU-Learning		&  56.54   &  23.91   &  24.63   &  5.32    &  11.96   &  7.13    &  6.81  	& 76.36 & 53.86 & 75.79		\\
		\hline
		HGL   &  \textbf{69.65}   &  \textbf{47.09}   &  \textbf{59.94}   &  \textbf{60.87}   &  \textbf{51.56}   &  \textbf{42.18}   &  \textbf{59.77}   & \textbf{84.45} & 67.51 & \textbf{77.56}     \\
		\hline
	\end{tabular}
	\caption{The F1 scores of Bert-CRF models with different denoising algorithms on ACE2005 test sets. We can see that the proposed HGL significantly outperforms other baselines, especially on those high noise-ratio settings.}
	\label{tab:bert_crf_f1}
\end{table*}

\subsection{Learning NER models with Denoised Data}
In this section, we investigate whether denoised data can effectively improve the training of a commonly-used NER model. Specifically, for each denoising algorithm, we regarded $N\times p$ weakly-labeled instances in the training set with the highest confidence derived from the denoiser as the golden entity mentions. Then the denoised data is used to train widely-used BERT-CRF NER taggers to evaluate the performance. The performance is evaluated using the final F1-score achieved in the original test set.

Table~\ref{tab:bert_crf_f1} shows the results. We can see that:

1) \textbf{Denoising is critical for distant supervision NER.} Compared with the naive distant supervision setting, using denoised data to train Bert-CRF tagger achieves significant improvements.

2) \textbf{HGL outperforms all other denoising baselines on all settings by a large margin, especially on those settings with larger noise rates.} This result is not surprising because, as we have shown before, the denoising performance of HGL is significantly better than other baselines. This again demonstrates the effectiveness of HGL in modeling noise distribution and instance-level confidence.

3) \textbf{There is still a gap between the supervised learning and distant supervision.} This is because even remarkable progress has been made by HGL, there still exists some errors in denoising. This indicates the potential improvement in denoising distant supervision NER in the future.

\begin{table}[!h]
	\centering
	\setlength{\belowcaptionskip}{-0.2cm}
		\begin{tabular}{l|rrr}
			\hline
			Method  & PER       & ORG       & GPE       \\
			\hline \hline
			Instance   &     74.79      &     71.70      &     69.85      \\
			XR         &     79.07      &     72.93      &     73.13      \\
			\hline
			HGL (False Positive Only)   &     78.88      &     72.69      &     76.06      \\
			\textbf{HGL + Mention Blocking}     &     \textbf{79.80}      &     \textbf{73.97}      &     \textbf{76.13}      \\
			\hline
		\end{tabular}
	\caption{The AUC scores of denoising performance on ACE2005\_NAM when false negative instances exist.}
	\label{tab:auc_neg}
\end{table}

\begin{figure}[!ht]
	\centering
	\setlength{\belowcaptionskip}{-0.3cm}
	\includegraphics[width=0.3\textwidth]{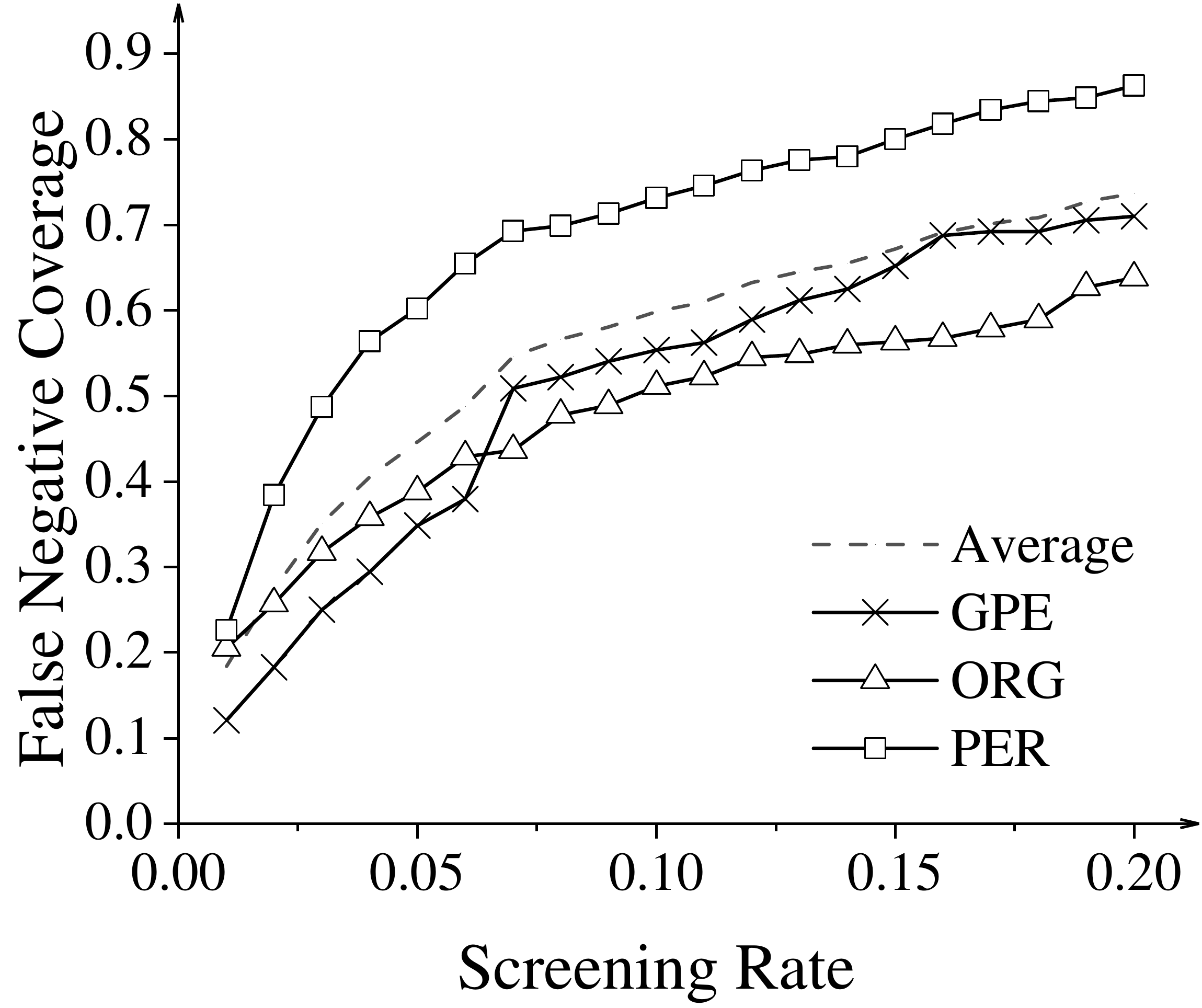}
	\caption{The coverage of false negative instances w.r.t. noun phrase screening rate. We can see that Mention Blocking can cover more than 50\% false negative instances by considering less than 10\% noun phrases.}
	\label{fig:men_cov}
\end{figure}

\subsection{Efficacy of Mention Blocking}

This section conducted experiments to verify the effectiveness of the proposed Mention Blocking approach and how well HGL can work when false negative exists. For this, we randomly dropped 50\% mentions in the dictionary derived from the ACE2005 training data. Therefore, the weakly-labelled data will contain a number of both false positive and false negative instances. We compared HGL with or without Mention Blocking to verify the effectiveness of processing false negative instances. We keep noun phrases that with top 10\% confidence in Mention Blocking as the potential false negative mentions, and introduce them into the HGL procedure.

Table~\ref{tab:auc_neg} shows the AUC scores of denoising performance under this setting~\footnote{The AUC scores does not consider the noun phrases that are not identified by either distant supervision or Mention Blocking, because all methods will treat them as negative instances and they will not influence the evaluation results.}. We can see that HGL with Mention Blocking significantly outperforms the one that only focus on false positive instances, as well as other baselines. This shows the effectiveness of the proposed Mention Blocking approach. To further demonstrate this, Figure~\ref{fig:men_cov} shows the coverage of the false negative instances with respect to the total noun phrases screened out. From this figure, we can see that Mention Blocking is very effective to find the potential entity mentions: top 10\% noun phrases screened out can cover more than 50\% false negative mentions. This verify the efficacy of Mention Blocking method.

\section{Experiments with Real-world Dictionaries}
To further verify the effectiveness of the proposed hypergeometric learning method on real-world DS-NER settings. We follow~\citet{peng-etal-2019-distantly} to further conduct an experiments on CoNLL2003 dataset, using a real-world dictionary as the source of distant supervision. 

\paragraph{Distant Supervision.}
We followed~\citet{peng-etal-2019-distantly} and used their extracted dictionary to conduct the evaluation. The dictionary was extracted from government websites~\footnote{http://www.ons.gov.uk/ons/index.html} and Wikipedia pages~\footnote{https://en.wikipedia.org/wiki/List\_of\_countries\_by\_natio
nal\_capital\_largest\_and\_second\-largest\_cities}. We then used the dictionaries to obtain weakly-labeled training instances using forward maximum matching. According to ~\citet{peng-etal-2019-distantly}, this dictionary is with very high precision but relatively low coverage rate on CoNLL03, and the denoising algorithm should be conducted to distinguish false negative instances. Therefore, we directly apply HGL on the negative instances.

\paragraph{Overall Performance.}

\begin{table}[!t]
	\centering
	\setlength{\belowcaptionskip}{-0.3cm}
	\resizebox{0.49\textwidth}{!}{
		\begin{tabular}{l|cccc|c}
			\hline 
					   			& PER		& ORG		& LOC		& MISC		& Overall      \\ \hline
			\hline
			XR  				& 91.97     & 64.46     & 71.00 	& 55.11      & 73.65    \\
			PU-Learning         & \textbf{93.93}     & 67.34     & 65.58 	& 50.94      & 72.34    \\
			\hline
			HGL  				& 92.78     & \textbf{68.30}     & \textbf{71.62} 	& \textbf{56.69}      & \textbf{74.87}    \\
			\hline
		\end{tabular}
	}
	\caption{DS-NER performance on CoNLL03 test set evaluated using span-based micro-F1 scores. We can see that HGL significantly outperform other denoising algorithms.}
	\label{tab:conll_overall}
\end{table}

\begin{table}[!t]
	\centering
	\setlength{\belowcaptionskip}{-0.3cm}
	\begin{tabular}{c|c|c|c|c|c}
		\hline
		Recall 				  & Model 		& PER 		& ORG 	& LOC 		& MISC 		\\ \hline \hline
		\multirow{3}{*}{25\%} & XR			& \textbf{99.00}  	& 71.41	& 85.38 	& 65.32 		\\ \cline{2-6} 
							  & PU-Learning	& 95.73  	& \textbf{87.22} & 86.45 	& 90.03		 \\ \cline{2-6} 
							  & HGL			& 96.73  	& 85.12 & \textbf{96.33} 	& \textbf{90.82}		 \\ \hline
		\multirow{3}{*}{50\%} & XR			& \textbf{98.32}  	& 60.20 & 82.25 	& 62.26 			\\ \cline{2-6} 
							  & PU-Learning	& 92.76  	& 77.03 & 75.90 	& \textbf{80.72} 	\\ \cline{2-6} 
							  & HGL			& 97.96  	& \textbf{78.22} & \textbf{90.29} 	& 66.68 	\\ \hline
		\multirow{3}{*}{75\%} & XR			& 96.70  	& 48.88 & 69.13 	& \textbf{45.32} 	\\ \cline{2-6} 
							  & PU-Learning	& 86.89  	& 55.88 & 41.78 	& 20.36 		\\ \cline{2-6} 
							  & HGL			& \textbf{97.93}  	& \textbf{70.40} & \textbf{70.98} 	& 43.04 	\\ \hline
	\end{tabular}
	\caption{The token-level precisions under different recall rates on the denoised distantly-supervised CoNLL03 training set. We can see that HGL achieved the best performance on the majority of settings.}
	\label{tab:conll_recall}
\end{table}

Table~\ref{tab:conll_overall} shows the overall experimental results measured using standard span-based micro-F1\footnote{From the source code provided by \citet{peng-etal-2019-distantly} on Github, we find that they used a token-based evaluation criteria, which is different from the most widely-used span-based criteria and therefore over-estimated NER performance. In this paper, we reproduced their results using the provided source code and compared it with HGL/XR using the standard span-based micrio-F1 metric.}. 
We can see that HGL significantly outperforms other baselines. Furthermore, Table~\ref{tab:conll_recall} shows the change of precision w.r.t. different target recall rates. We can see that under the majority of experiment groups, HGL can achieve better precision than other previous methods. This demonstrate that HGL can achieve steadily better performance under ranking measurements, i.e., instances obtained high confidence from HGL are more likely to be true entity mentions. All these experimental results consist with results in the previous section, which further demonstrate the effectiveness of HGL.

\section{Conclusions}
This paper proposes a new denoising algorithm -- Hypergeometric Learning, which can model and exploit both the instance-level evidence and the underlying noise distribution for denoising DS-NER. Experimental results on both high-precision and high-recall DS-NER settings show that, by further taking the corpus-level noise distribution into consideration, HGL can effectively denoise the DS-labeled data, learn an effective NER model, and achieve robust performance on different entity types and different dataset settings.

\newpage

\section{Acknowledgements}
This research work is supported by National Key Research and Development Program of China under Grant No.2019YFC1521200, the National Natural Science Foundation of China under Grants no. U1936207 and 61772505, Beijing Academy of Artificial Intelligence (BAAI2019QN0502), scientific research projects of the State Language Commission (YW135-78), and in part by the Youth Innovation Promotion Association CAS(2018141). Moreover, we thank all reviewers for their valuable comments and suggestions.

\bigskip

\bibliography{aaai21}

\end{document}